\documentclass[letterpaper]{article} 
\usepackage[preprint]{aaai2027} 
\usepackage[hyphens]{url} 
\usepackage{graphicx} 
\usepackage{natbib} 
\usepackage{caption} 
\usepackage{booktabs}

\usepackage{amsmath}
\usepackage{amssymb}
\usepackage{multirow}

\newcommand{\method}{GeoFF3D}
\newcommand{\framework}{SLRF}
\newcommand{\best}[1]{\textbf{#1}}

\title{GeoFF3D: Coordinate-Anchored Feed-Forward Reconstruction for Large-Scale UAV Mapping}
\author{
Xiang Yang\textsuperscript{1},
Yongli Wang\textsuperscript{1},
Yunsheng Zhang\textsuperscript{1,2,*},
Jun Li\textsuperscript{3},
Hao Chen\textsuperscript{3},
Haifeng Li\textsuperscript{1}
}
\affiliations{
\textsuperscript{1}School of Geosciences and Info-Physics, Central South University, Changsha, China\\
\textsuperscript{2}Hunan Engineering Research Center of 3D Real Scene Construction and Application Technology, Changsha, China\\
\textsuperscript{3}College of Electronic Science and Technology, National University of Defense Technology, Changsha, China\\
\textsuperscript{*}Corresponding author. Email: \texttt{zhangys@csu.edu.cn}
}

\begin{document}
\maketitle
\begin{abstract}
Existing feed-forward 3D reconstruction methods typically process a bounded number of images and recover cameras and geometry in local or internally normalized frames. Extending them to large-scale UAV mapping requires scalable multi-chunk processing and reliable aggregation, while full Sim(3) alignment can become unstable for near-collinear trajectories. We present GeoFF3D, which combines a coordinate-anchored model with a spatial large-scale reconstruction framework (SLRF). The model uses georeferenced camera translations and optional geometric priors to predict camera poses and dense point maps directly in a gravity-aligned Z-up metric frame. SLRF partitions images into spatially overlapping chunks, propagates shared-view priors, and aggregates local reconstructions hierarchically, while remaining applicable to different bounded-view models. Across nine aerial mapping blocks, GeoFF3D achieves the best average reconstruction quality, improving F@5 from 0.829 for Pi3X + SLRF to 0.877. On long UAVScenes sequences, it reaches 0.848, compared with 0.687 for Pi3X + SLRF and 0.451 for the strongest evaluated SLAM/streaming baseline. GeoFF3D reconstructs 2,000 images in approximately five minutes, demonstrating scalable and robust large-scale UAV reconstruction. The code is available at \url{https://github.com/yanxian-ll/GeoFF3D}.
\end{abstract}


\section{Introduction}\label{sec:introduction}

Unmanned aerial vehicle (UAV) oblique photogrammetry is widely used for large-scale 3D scene acquisition in
urban modeling, terrain surveying, disaster assessment, and infrastructure inspection. Conventional
photogrammetric systems recover camera poses and scene geometry through multi-stage geometric estimation and
iterative optimization, providing reliable mapping products at the cost of substantial computation
\cite{schonberger2016sfm,schonberger2016mvs}. Recent feed-forward reconstruction models instead directly
predict camera parameters, depth maps, or dense point maps from multi-view images, offering a more efficient
alternative \cite{wang2024dust3r,leroy2024mast3r,yang2025fast3r,wang2025vggt,wang2025pi3}. However, most
existing models process only a bounded number of images and reconstruct them in reference-camera frames or
internally normalized scene representations, whereas practical UAV surveys contain hundreds or thousands of
views and require outputs in a unified georeferenced, gravity-aligned metric frame.

\begin{figure}[t]
  \centering
  \includegraphics[width=\columnwidth]{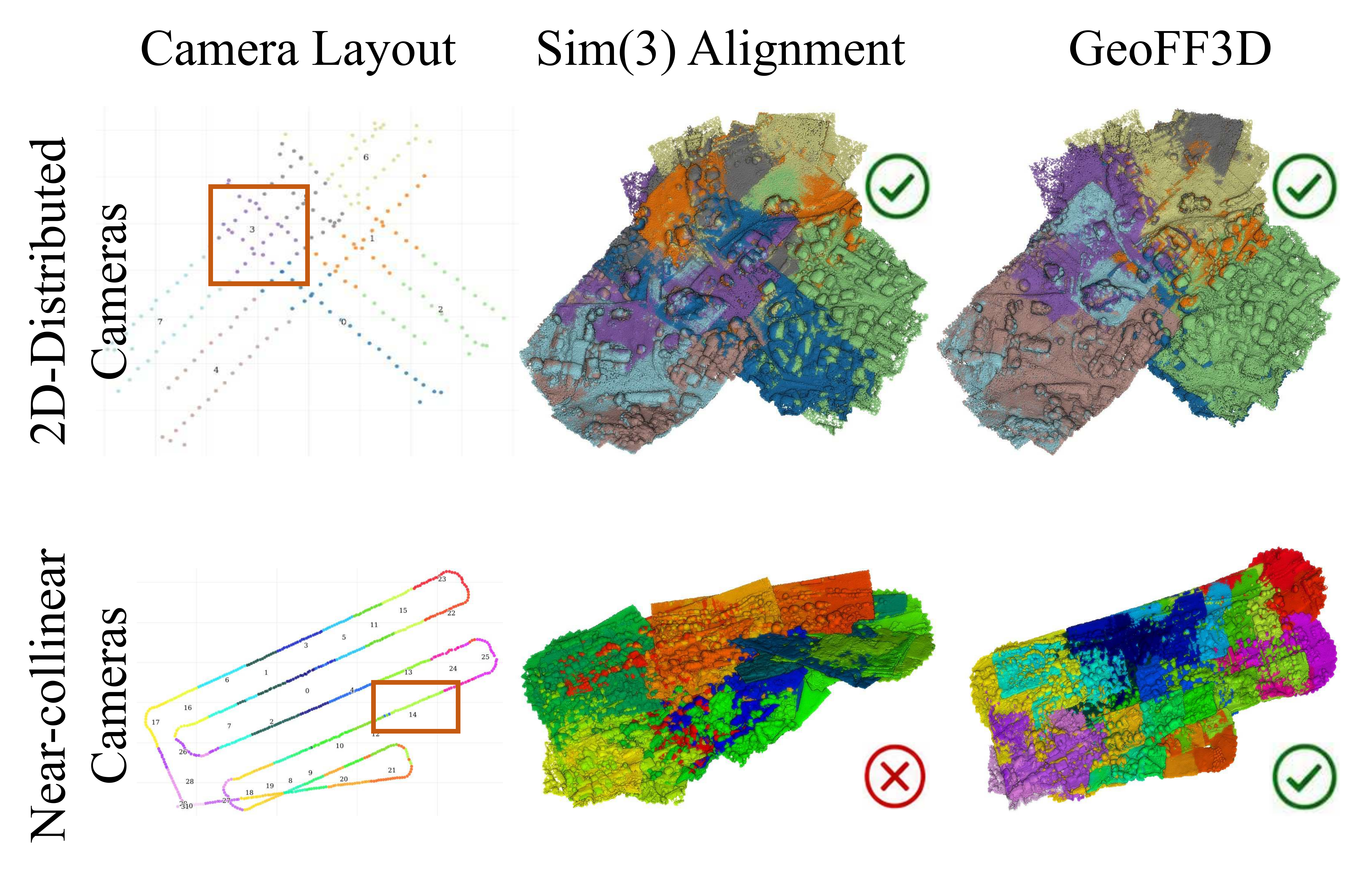}
  \caption{Effect of camera layout on post-hoc alignment. Full Sim(3) is well constrained for
    two-dimensionally distributed cameras but may tilt a near-collinear reconstruction, whereas GeoFF3D
    preserves the gravity-aligned Z-up orientation.}
  \label{fig:camera_layout_alignment}
\end{figure}

\begin{figure*}[t!]
  \centering
  \includegraphics[width=1.0\textwidth]{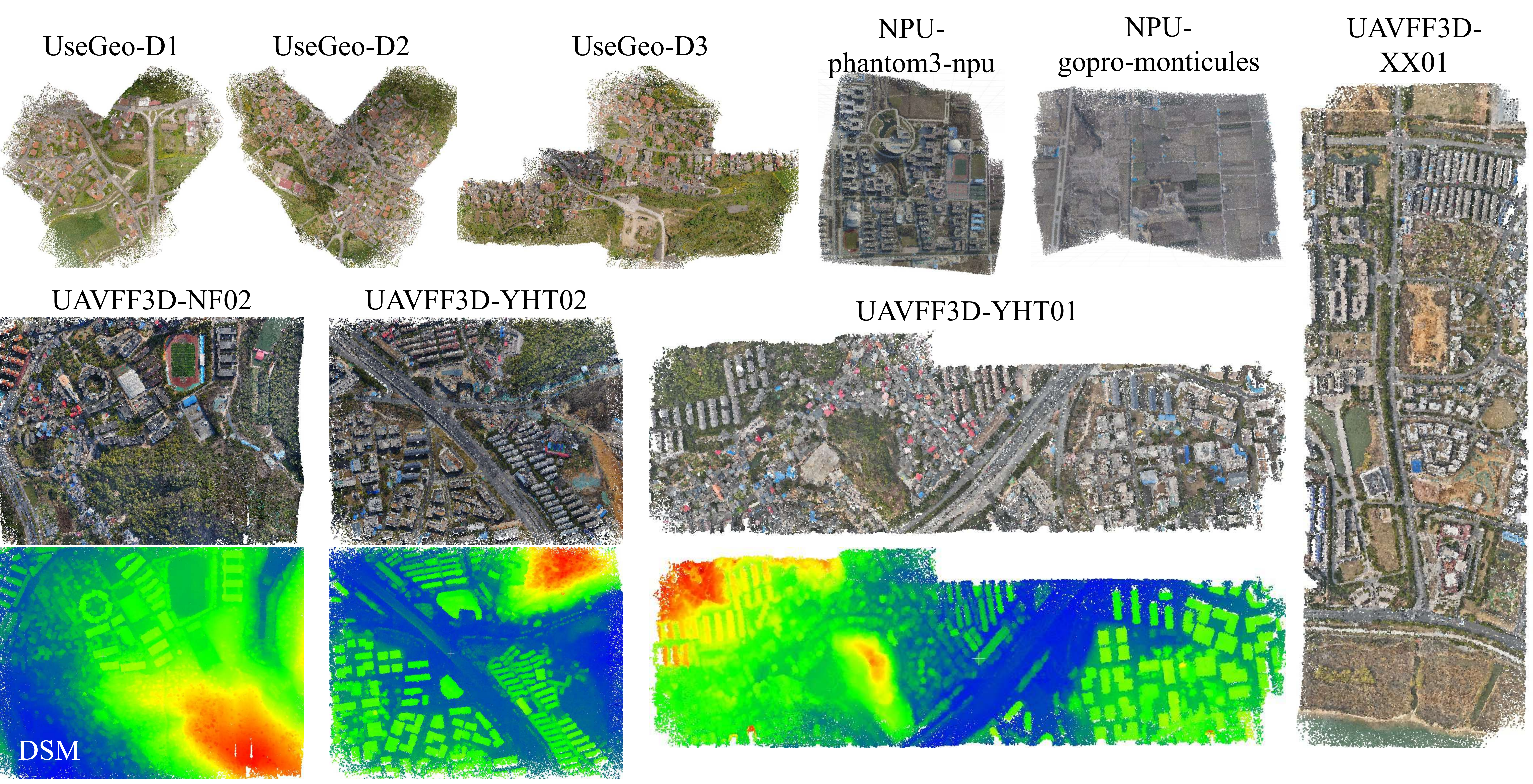}
  \caption{Qualitative reconstruction results of GeoFF3D across diverse aerial mapping scenes from UseGeo,
    NPU-DroneMap, and UAVFF3D-Real. The results cover multi-strip surveys with varying spatial extents, camera
    platforms, and scene structures.}
  \label{fig:qualitative_gallery}
\end{figure*}

This gap introduces two coupled challenges. First, georeferencing a locally predicted reconstruction commonly
requires estimating a full Sim(3) transformation between predicted camera centers and positioning priors. As
illustrated in Fig.~\ref{fig:camera_layout_alignment}, this transformation is well constrained for
two-dimensionally distributed cameras, but may become unstable for near-collinear UAV trajectories. In such
cases, roll and pitch are weakly constrained, so the dense reconstruction may be incorrectly tilted even when
the aligned camera centers appear accurate. Second, large image collections must be divided into overlapping
chunks. Because each chunk is reconstructed independently, neighboring chunks may disagree in their coordinate
frames and in the depths predicted for shared images, producing duplicated surfaces, vertical offsets, and
visible seams. These difficulties are further amplified in multi-strip UAV surveys, where temporal proximity
does not necessarily indicate spatial overlap.

Our key insight is to use georeferencing priors to define the prediction coordinate frame itself, rather than
reconstructing in an internal frame and registering the result afterward. Based on this insight, we present
GeoFF3D, which consists of a coordinate-anchored model and a spatial large-scale reconstruction framework
(SLRF). The coordinate-anchored model uses georeferenced translations and optional geometric priors to
directly predict camera poses and dense point maps in a gravity-aligned Z-up metric frame. The translation
anchors provide approximate metric scale and spatial placement, while the Z-up representation stabilizes the
vertical direction. SLRF partitions images into spatially overlapping chunks, processes them from the scene
center outward, propagates shared-view priors when supported, and hierarchically aggregates local
reconstructions using gravity-preserving transformations. Together, these designs improve cross-chunk consistency. SLRF can wrap
different bounded-view feed-forward predictors while enabling prior propagation only when supported.

We evaluate GeoFF3D on multi-strip aerial blocks and long UAV sequences. GeoFF3D achieves the best average performance across nine aerial
blocks and a substantially larger margin on long UAVScenes sequences. These results show that
coordinate-anchored prediction improves aggregation across general multi-strip surveys and is particularly
beneficial for long, near-collinear trajectories, while SLRF scales to collections containing thousands of
images. Our main contributions are:

\begin{enumerate}
  \item We introduce a coordinate-anchored feed-forward model that uses approximate georeferenced camera
        translations and optional geometric priors to directly recover camera poses and dense geometry in a
        gravity-aligned Z-up metric frame, reducing reliance on unstable full Sim(3) post-hoc alignment.

  \item We introduce SLRF, which combines footprint-guided spatial chunking, center-outward prior propagation, and gravity-preserving hierarchical aggregation. It can be instantiated with different bounded-view predictors and scales to UAV collections containing hundreds to thousands of images.

\end{enumerate}


\section{Related Work}\label{sec:related_work}

\paragraph{Feed-forward 3D reconstruction.}
Existing feed-forward methods typically recover cameras and geometry in reference-camera frames or internally
normalized scene representations. Methods such as DUSt3R, MASt3R, and VGGT anchor their outputs to a selected
input camera \cite{wang2024dust3r,leroy2024mast3r,yang2025fast3r,wang2025vggt,wang2026vggtomega}. $\pi^3$
instead removes the designated reference view and predicts permutation-equivariant cameras and geometry, while
retaining an internally normalized representation \cite{wang2025pi3,wang2024spann3r,zhang2025flare}.
Prior-conditioned methods such as Pow3R, MapAnything, and MASt3R-Fusion further incorporate camera or
geometric observations to guide reconstruction or downstream optimization
\cite{jang2025pow3r,keetha2025mapanything,lu2025matrix3d,khafizov2025gcut3r,liu2025worldmirror,
  zhou2025mast3rfusion}.
In contrast, GeoFF3D uses georeferenced camera translations to define the prediction frame itself, directly
recovering cameras and dense geometry in a gravity-aligned metric frame.

\paragraph{Scalable feed-forward reconstruction.}
Recent methods extend feed-forward reconstruction to large image collections and streaming inputs using spatial memories, persistent states, causal attention, token compression, or test-time adaptation
\cite{wang2024spann3r,wang2025cut3r,chen2025long3r,lan2025stream3r,zhuo2025streamvggt,li2025wint3r,
  yuan2026infinitevggt,su2026xstreamvggt,zou2026retrievevggt,robbyant2026lingbotmap,chen2025ttt3r,
  jin2026zipmap,zhang2026loger,xie2026scal3r,shen2025fastvggt}.
Other systems reconstruct overlapping clips or submaps and enforce global consistency through learned
registration, loop closure, pose graphs, or geometric optimization
\cite{liu2025slam3r,deng2025vggtlong,maggio2025vggtslam,maggio2026vggtslam2,murai2024mast3rslam,
  zhang2025vistaslam,zhang2025talo,li2025sing3rslam,li2024megasam,huang2025vipe}.
These approaches are primarily designed for temporally ordered videos or sequentially neighboring submaps and
commonly rely on registration or optimization to enforce global consistency. In contrast, SLRF partitions UAV
images according to spatial coverage rather than acquisition order, enabling GeoFF3D to handle both
multi-strip aerial blocks and long sequences without additional global optimization.


\section{Method}\label{sec:method}


\subsection{Problem Formulation}\label{subsec:problem_formulation}

Given a collection of \(N\) UAV images, where \(N\) may reach hundreds or thousands, a feed-forward
model can process at most \(M\) views in one pass, with \(N\gg M\). Approximate georeferenced camera
translations provide the primary coordinate anchors, while camera rotations, intrinsics, and depth
observations are optional. Our goal is to recover all camera poses and dense geometry in a shared
georeferenced Z-up metric frame.

The complete reconstruction is formulated as
\[
\begin{aligned}
\left\{\left(\widehat{\mathbf T}_i^w,\widehat{\mathbf X}_i^w\right)\right\}_{i=1}^{N}
&=
\mathcal A_{\mathcal T}\!\left(
\left\{f_\theta\!\left(\mathcal D_{\mathcal S_k},\mathcal Q_k\right)\right\}_{k=1}^{K}
\right), \\
\left\{\mathcal S_k\right\}_{k=1}^{K}
&=
\mathcal P_{\mathrm{foot}}(\mathcal D),
\qquad
|\mathcal S_k|\le M\ll N,
\end{aligned}
\]
where \(\mathcal P_{\mathrm{foot}}\) partitions the image collection \(\mathcal D\) into spatially overlapping
chunks, \(f_\theta\) predicts each chunk using its georeferencing inputs and propagated priors
\(\mathcal Q_k\), and \(\mathcal A_{\mathcal T}\) aggregates the chunk predictions into globally consistent
world-frame cameras \(\widehat{\mathbf T}_i^w\) and dense point maps \(\widehat{\mathbf X}_i^w\).

\subsection{Coordinate-Anchored Feed-Forward Reconstruction}\label{subsec:coordinate_anchored_reconstruction}

\begin{figure}[t]
  \centering
  \includegraphics[width=1.0\columnwidth]{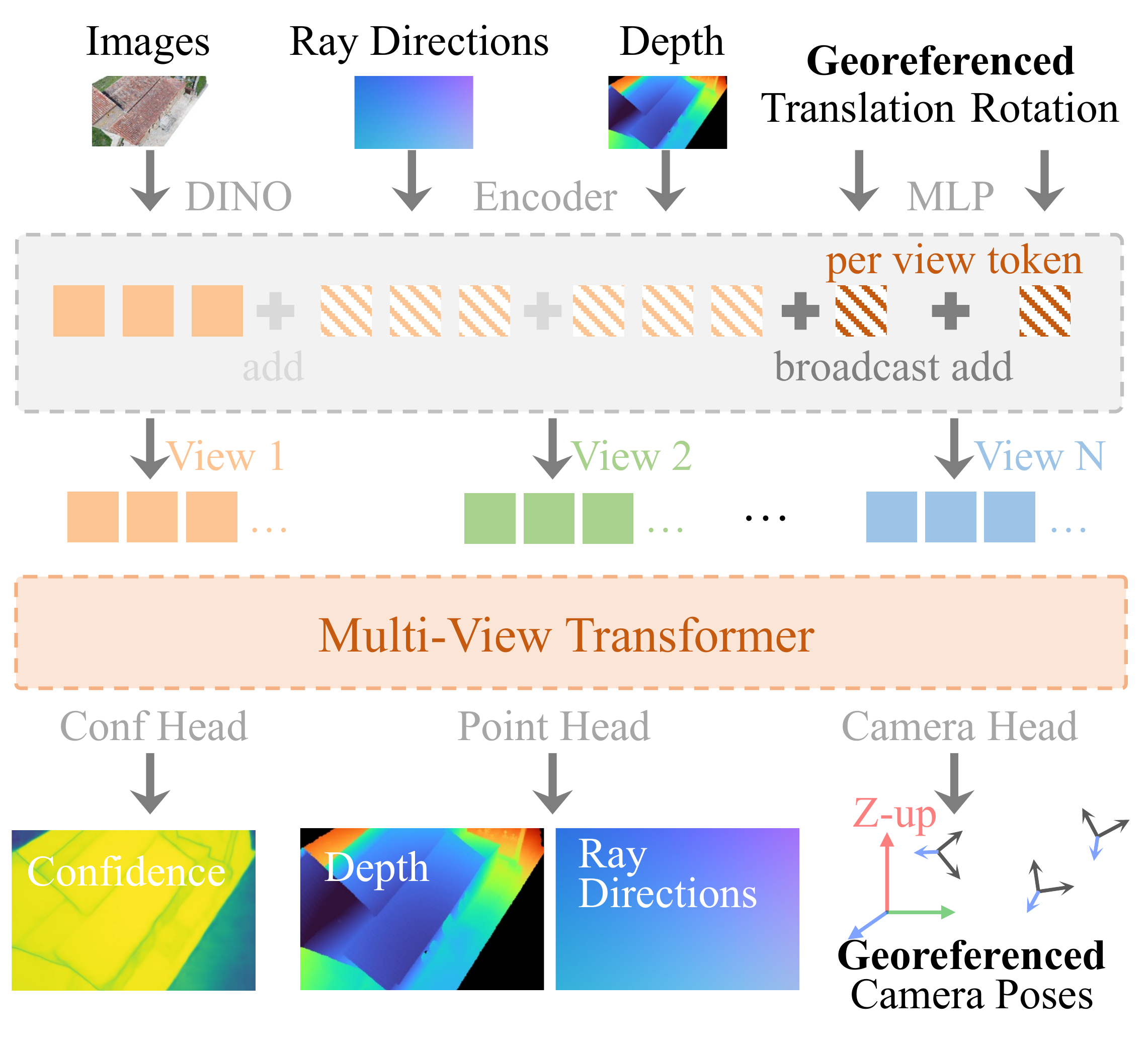}
  \caption{Coordinate-anchored feed-forward model.}
  \label{fig:coordinate_anchored_model}
\end{figure}

The first component of GeoFF3D is a coordinate-anchored feed-forward model for bounded-view reconstruction.
Given a spatial chunk containing at most \(M\) views, the model fuses image features with available
geometric priors and predicts cameras and dense geometry in the shared georeferenced Z-up world frame.
Figure~\ref{fig:coordinate_anchored_model} summarizes this component.

\begin{figure*}[t!]
  \centering
  \includegraphics[width=1.0\textwidth]{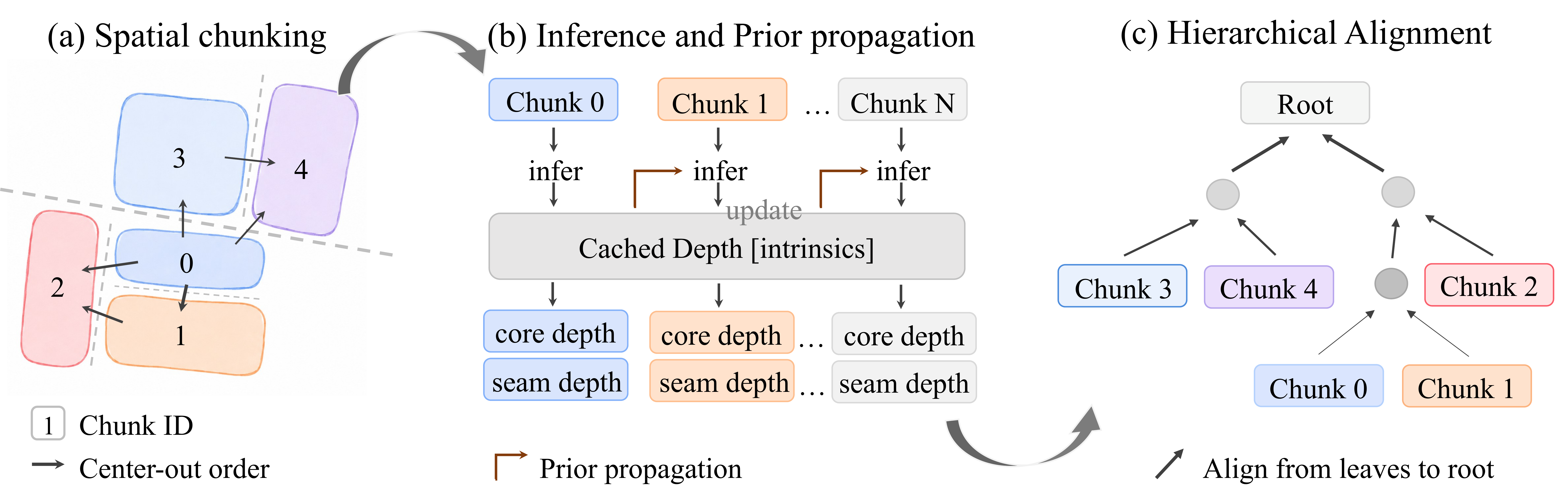}
  \caption{The spatial large-scale reconstruction framework (SLRF) of GeoFF3D. Images are first partitioned
    into overlapping core and seam views according to their estimated ground footprints. The resulting chunks
    are processed from the scene center outward, while cached depth and camera intrinsics from shared views
    are propagated to subsequent chunks. The local reconstructions are finally merged from leaves to root
    using gravity-preserving alignment.}
  \label{fig:slrf_overview}
\end{figure*}

\subsubsection{Coordinate Frame}\label{subsubsec:coordinate_frame}
The coordinate-anchored model predicts cameras and geometry in a local georeferenced Cartesian frame, whose
Z-axis points opposite to gravity and whose X--Y plane is parallel to the local horizontal plane. Approximate
camera translations are expressed in this frame and provide metric scale and spatial placement. For numerical
stability, we compute a translation center $\mathbf c_T$ and scene scale $s_T$ from the available
translation anchors, and use them to normalize the camera translations and metric depth priors within each
chunk. After inference, the predicted cameras and geometry are restored to the georeferenced metric frame
using $\mathbf c_T$ and $s_T$. Consequently, independently processed chunks are predicted in a shared
Z-up frame before aggregation.

\subsubsection{Prior Encoding}\label{subsubsec:prior_encoding}

For view \(i\), the image encoder produces visual patch tokens \(F_i^I\). Ray directions and depth
observations are encoded as patch-level features \(F_i^R\) and \(F_i^D\), preserving their spatial
correspondence with the image tokens. In contrast, normalized georeferenced translations and optional
georeferenced rotations are encoded as view-level embeddings \(e_i^t\) and \(e_i^r\), which are broadcast to
all patch tokens of the corresponding view. The fused representation is
\[
F_i = F_i^I + m_i^{\mathrm{ray}}F_i^R + m_i^D F_i^D + m_i^t e_i^t + m_i^r e_i^r,
\]
where the binary masks indicate the availability of each prior. During training, prior dropout randomly
removes individual inputs, allowing the same model to handle complete, partial, or sparse geometric
observations. The fused tokens are processed by alternating intra-view and inter-view attention before being
passed to the prediction heads. Translation priors serve as coordinate-conditioning signals rather than being
directly copied to the output.

\subsubsection{Prediction}\label{subsubsec:world_frame_prediction}

The camera head predicts a complete camera-to-world transformation
\(\widehat{\mathbf W}_i^{c\rightarrow w}\in SE(3)\) for each view. In parallel, the dense head predicts an
image-plane ray \(\widehat{\mathbf r}_i(u)\) and a positive Z-depth \(\widehat d_i(u)\) for each pixel
\(u\). Their product defines the camera-frame point, which is transformed and de-normalized into the
world frame:
\[
\widehat{\mathbf X}_i^w(u)
=
s_T\,\pi_3\!\left(
\widehat{\mathbf W}_i^{c\rightarrow w}
\begin{bmatrix}
\widehat d_i(u)\widehat{\mathbf r}_i(u) \\
1
\end{bmatrix}
\right)
+
\mathbf c_T.
\]
Here, \(\pi_3(\cdot)\) extracts the three-dimensional coordinates from the homogeneous representation. A
separate head predicts per-pixel confidence for suppressing unreliable geometry during fusion.

\subsubsection{Training}\label{subsubsec:training}

We initialize the coordinate-anchored model of GeoFF3D from a pretrained Pi3X checkpoint and train it on
UAVFF3D~\cite{yang2026uavff3d} and BlendedMVS~\cite{yao2020blendedmvs}. All scenes are converted to a common
gravity-aligned Z-up coordinate system. The overall training objective is
\[
\mathcal L
=
\lambda_{\mathrm{local}}\mathcal L_{\mathrm{local}}
+
\lambda_{\mathrm{world}}\mathcal L_{\mathrm{world}}
+
\lambda_{\mathrm{pose}}\mathcal L_{\mathrm{pose}}
+
\lambda_{\mathrm{grav}}\mathcal L_{\mathrm{grav}},
\]
where we set \(\lambda_{\mathrm{local}}=0.5\), \(\lambda_{\mathrm{world}}=1.0\),
\(\lambda_{\mathrm{pose}}=0.2\), and \(\lambda_{\mathrm{grav}}=1.0\). Following Pi3~\cite{wang2025pi3},
\(\mathcal L_{\mathrm{local}}\) supervises camera-frame point maps, Z-depths, image-plane rays, and relative
camera poses. The world-frame point and
pose losses supervise dense geometry and absolute camera poses in the normalized georeferenced frame, enabling
the model to learn the coordinate system defined by the translation anchors rather than only reconstructing an
internally normalized scene. The gravity loss minimizes the cosine discrepancy between the predicted and
ground-truth camera up directions. It primarily constrains roll and pitch, thereby further stabilizing camera
orientation in the anchor-defined Z-up frame.

Training proceeds in two stages. We first train for 80 epochs with a base resolution of 224 pixels to
establish coordinate-anchored Z-up prediction. We then fine-tune the model for 10 epochs at 518 pixels, while
randomly dropping and perturbing the geometric priors to improve robustness to incomplete and noisy
georeferencing inputs.

\begin{figure*}[t!]
  \centering
  \includegraphics[width=1.0\textwidth]{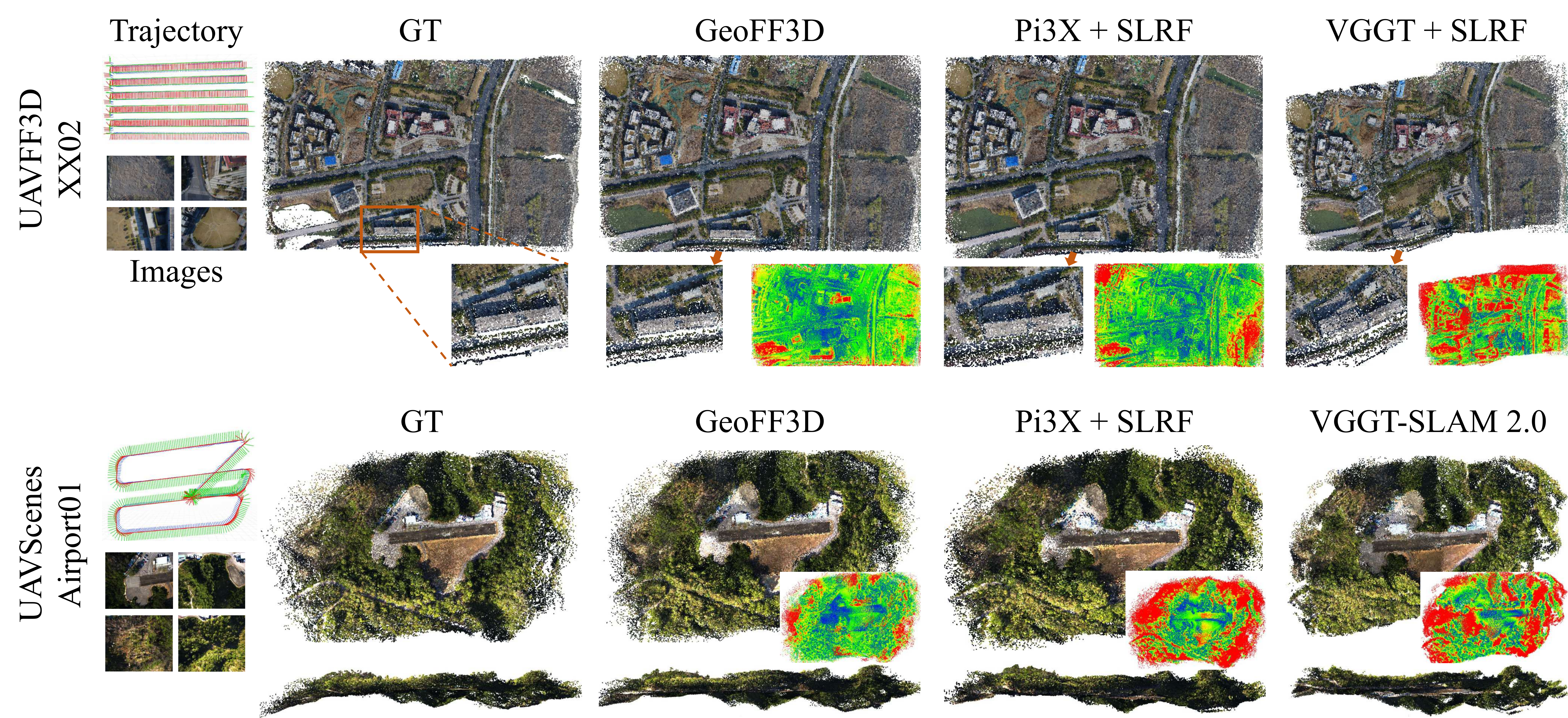}
  \caption{Qualitative comparison on UAVFF3D-Real XX-02 (top) and the elongated UAVScenes Airport01 sequence
    (bottom). Main panels show top views, while the auxiliary panels visualize local structure,
    point-to-ground-truth distance, and side profiles. The methods exhibit different boundary, duplication,
    and vertical-profile errors.}
  \label{fig:qualitative_comparison}
\end{figure*}

\subsection{Spatial Large-Scale Reconstruction Framework}\label{subsec:slrf}

Large UAV collections exceed the single-pass view budget and must be processed as multiple local chunks. The
second component of GeoFF3D is its spatial large-scale reconstruction framework (SLRF), which wraps the
bounded-view coordinate-anchored model with footprint-guided chunking, center-outward processing, and
hierarchical aggregation, as shown in Fig.~\ref{fig:slrf_overview}. 

\subsubsection{Footprint-Guided Spatial Chunking}
\label{subsubsec:footprint_guided_chunking}

We first estimate the ground footprint of each image from available georeferencing information, including GNSS positions and IMU orientations. Based on these footprints, we recursively partition the image collection into spatially compact leaf chunks rather than following acquisition order. We build an adaptive binary tree by splitting
the root along the along-track axis and each subsequent node at the
median along the axis of largest center spread, until every leaf contains
at most \(\lfloor0.7M\rfloor\) core views. Each leaf \(k\) contains
a core-view set \(\mathcal C_k\) and additional seam views
\(\mathcal O_k\) selected from neighboring regions:
\[
\mathcal S_k=\mathcal C_k\cup\mathcal O_k,
\qquad
|\mathcal C_k|\leq\lfloor0.7M\rfloor,
\qquad
|\mathcal S_k|\leq M.
\]
Each image belongs to one core set but may also appear as a seam view in adjacent chunks. Core-view geometry
is retained for final fusion, while seam views provide overlap for depth propagation and cross-chunk
alignment.

\subsubsection{Center-Outward Inference and Prior Propagation}\label{subsubsec:center_outward_inference}

We construct a spatial adjacency graph over the leaf chunks, select the chunk closest to the scene center, and
process the chunks in breadth-first order from the center outward. We maintain a depth cache indexed by image
identity. Before inferring each chunk, cached depths of its shared views are retrieved and supplied to the
model as input depth priors. After inference, low-confidence regions are removed using a confidence quantile
threshold of 0.25. The filtered depths are then used to update the cache, with core-view predictions preferred
when duplicate estimates exist. The updated cache is used for subsequent chunks. When camera intrinsics are
unavailable and all images are captured by the same camera, the average intrinsics predicted from the first
chunk are used as input priors for all subsequent chunks. Reusing shared-view priors reduces depth
discrepancies for the same images across neighboring chunks and improves cross-chunk geometric consistency.

\subsubsection{Gravity-Aligned Hierarchical Aggregation}\label{subsubsec:gravity_aligned_aggregation}

Although all chunks are predicted in the common georeferenced Z-up world frame, residual errors may remain
because of noisy georeferencing and local prediction uncertainty. We first anchor each leaf chunk to its
georeferenced translation priors using a gravity-aligned similarity transformation (GA-Sim)
\[
\mathcal T_{\mathrm{GA\text{-}Sim}}(\mathbf x)
=
s\mathbf R_z(\theta)\mathbf x+\mathbf t,
\]
which estimates scale, yaw, and three-dimensional translation while preserving roll and pitch. Sibling nodes
are then refined using a gravity-aligned rigid transformation (GA-Rigid)
\[
\mathcal T_{\mathrm{GA\text{-}Rigid}}(\mathbf x)
=
\mathbf R_z(\Delta\theta)\mathbf x+\Delta\mathbf t,
\]
which estimates only residual yaw and translation from shared-view correspondences. The alignment proceeds
bottom-up along the footprint tree until a single root reconstruction is obtained. The accumulated
transformations are finally applied to all leaf chunks, and only confidence-filtered core-view geometry is
fused into the final point cloud. These gravity-preserving transformations correct residual cross-chunk
misalignment without disturbing the shared Z-up orientation.

\begin{table*}[t!]
  \centering
  {
    \renewcommand{\arraystretch}{1.08}
    \setlength{\tabcolsep}{2.0pt}
    \begin{tabular}{@{}llcccc|cccc|cccc@{}}
      \toprule
      \multirow{2}{*}{Dataset}
      & \multirow{2}{*}{Scene}
      & \multicolumn{4}{c|}{VGGT + SLRF}
      & \multicolumn{4}{c|}{Pi3X + SLRF}
      & \multicolumn{4}{c}{GeoFF3D} \\
      & & Acc.$\downarrow$ & Comp.$\downarrow$ & F@1$\uparrow$ & F@5$\uparrow$
      & Acc.$\downarrow$ & Comp.$\downarrow$ & F@1$\uparrow$ & F@5$\uparrow$
      & Acc.$\downarrow$ & Comp.$\downarrow$ & F@1$\uparrow$ & F@5$\uparrow$ \\
      \midrule
      \multirow{6}{*}{UAVFF3D}
      & YHT01
      & 4.86 & \underline{3.68} & 0.084 & 0.688
      & \underline{4.21} & 3.87 & \underline{0.096} & \underline{0.720}
      & \textbf{3.00} & \textbf{2.44} & \textbf{0.154} & \textbf{0.880} \\
      & YHT02
      & 5.36 & 5.19 & 0.086 & 0.658
      & \textbf{2.94} & \textbf{2.89} & \textbf{0.128} & \textbf{0.860}
      & \underline{3.19} & \underline{3.25} & \underline{0.087} & \underline{0.834} \\
      & NF01
      & 8.19 & 6.16 & 0.027 & 0.486
      & \textbf{4.12} & \textbf{3.32} & \textbf{0.057} & \textbf{0.771}
      & \underline{4.15} & \underline{3.34} & \underline{0.054} & \underline{0.766} \\
      & NF02
      & \underline{5.37} & \textbf{5.05} & 0.069 & \textbf{0.651}
      & 6.55 & 6.93 & \underline{0.070} & 0.588
      & \textbf{5.19} & \underline{6.25} & \textbf{0.076} & \underline{0.596} \\
      & XX01
      & 8.81 & 7.13 & 0.035 & 0.476
      & \underline{3.92} & \underline{3.40} & \underline{0.113} & \underline{0.793}
      & \textbf{2.65} & \textbf{2.16} & \textbf{0.149} & \textbf{0.939} \\
      & XX02
      & 2.70 & \underline{2.31} & 0.220 & 0.896
      & \textbf{2.02} & \textbf{1.90} & \textbf{0.270} & \textbf{0.951}
      & \underline{2.17} & \underline{2.30} & \underline{0.255} & \underline{0.935} \\
      \midrule
      \multirow{3}{*}{UseGeo}
      & D1
      & 1.65 & 1.17 & 0.512 & 0.970
      & \underline{1.29} & \underline{1.11} & \underline{0.575} & \underline{0.984}
      & \textbf{1.15} & \textbf{1.02} & \textbf{0.639} & \textbf{0.989} \\
      & D2
      & 3.62 & 3.18 & \underline{0.329} & 0.830
      & \underline{3.02} & \underline{2.79} & 0.324 & \underline{0.850}
      & \textbf{1.51} & \textbf{1.30} & \textbf{0.471} & \textbf{0.978} \\
      & D3
      & 2.56 & 1.76 & 0.322 & 0.917
      & \underline{2.00} & \underline{1.41} & \underline{0.440} & \underline{0.945}
      & \textbf{1.42} & \textbf{1.21} & \textbf{0.517} & \textbf{0.981} \\
      \midrule
      Overall
      & Average
      & 4.79 & 3.96 & 0.187 & 0.730
      & \underline{3.34} & \underline{3.07} & \underline{0.230} & \underline{0.829}
      & \textbf{2.72} & \textbf{2.59} & \textbf{0.267} & \textbf{0.877} \\
      \bottomrule
    \end{tabular}
  }
  \caption{Quantitative comparison on aerial mapping blocks.}
  \label{tab:aerial_block_results}
\end{table*}

\section{Experiments}\label{sec:experiments}

\subsection{Experimental Setup}\label{subsec:experimental_setup}

\paragraph{Datasets.}
We evaluate GeoFF3D on four UAV datasets with complementary roles. We use all three
UseGeo~\cite{nex2024usegeo} blocks, containing 224--327 images per scene, and six
UAVFF3D-Real~\cite{yang2026uavff3d} blocks, containing 387--1,177 images per scene, for quantitative
aerial-block evaluation. The six UAVFF3D-Real test scenes are geographically disjoint from the training
set; no image, flight, or physical site is used for training or model selection.
For long-sequence evaluation, we select eight UAVScenes~\cite{wang2025uavscenes} sequences with 1,317--2,589
original frames and uniformly sample them with a stride of 3 for all methods. We additionally use all 12
NPU-DroneMap~\cite{bu2016map2dfusion} keyframe sequences, containing 285--648 images per scene, for
qualitative evaluation because complete dense reference geometry is unavailable.

\paragraph{Baselines.}
For aerial blocks, we compare GeoFF3D with VGGT + SLRF and Pi3X + SLRF using UAVFF3D-fine-tuned
checkpoints~\cite{yang2026uavff3d}. VGGT receives no pose priors; translations are used only by SLRF for
footprint construction and alignment. Pi3X and GeoFF3D receive the same perturbed translation and rotation
priors. VGGT + SLRF and Pi3X + SLRF use full Sim(3) for leaf and inter-chunk alignment, whereas GeoFF3D
uses GA-Sim and GA-Rigid. Predicted depth is propagated only when supported. For long sequences, we additionally compare with VGGT-SLAM (Sim(3)), VGGT-SLAM 2.0, VGGT-Long, TTT3R, and
LingBot-Map
\cite{maggio2025vggtslam,maggio2026vggtslam2,deng2025vggtlong,chen2025ttt3r,robbyant2026lingbotmap}.

\paragraph{Metrics.}
We report Accuracy, Completeness, F@1, and F@5 for datasets with reference geometry, where F@1 and F@5 denote
the F-scores at 1~m and 5~m distance thresholds, respectively. All reported averages are computed over scenes.
All methods receive the same camera-center-based global Sim(3) and
point-cloud ICP for aligned geometry evaluation. This is separate from the per-chunk reconstruction protocols.

\paragraph{Implementation details.}
All methods process images with a maximum side length of 518 pixels. GeoFF3D, VGGT + SLRF, and Pi3X + SLRF use
a chunk budget of 30 views. The remaining methods use their released checkpoints and recommended inference
configurations. We train GeoFF3D using two NVIDIA A100 GPUs with 40~GB of memory each and evaluate all methods
using a single NVIDIA A100 GPU with 40~GB of memory. At test time, Pi3X + SLRF and GeoFF3D share pose priors generated by perturbing ground-truth translations
and rotations. Gaussian noise is added to horizontal translation, vertical translation, and yaw with standard
deviations of $0.5$~m, $0.8$~m, and $1.0^\circ$ and limits of $2.0$~m, $2.0$~m, and $3.0^\circ$.


\subsection{Large-Scale UAV Reconstruction}\label{subsec:large_scale_uav_reconstruction}

We evaluate GeoFF3D on multi-strip aerial mapping blocks and long UAV sequences.
Table~\ref{tab:aerial_block_results} reports the per-block results, Table~\ref{tab:uavscenes_results}
summarizes the average performance on UAVScenes, and Fig.~\ref{fig:qualitative_comparison} provides
representative qualitative comparisons. Figure~\ref{fig:qualitative_gallery} further presents GeoFF3D
reconstructions across aerial mapping scenes with different spatial extents and acquisition platforms.

\paragraph{Aerial mapping blocks.}
Across the nine UseGeo and UAVFF3D-Real blocks, GeoFF3D obtains the
best average Accuracy, Completeness, F@1, and F@5, with 2.72~m,
2.59~m, 0.267, and 0.877, respectively. Compared with Pi3X + SLRF,
it reduces Accuracy and Completeness by 18.7\% and 15.8\%, while
improving F@1 and F@5 by 0.036 and 0.048. The gains at both thresholds
indicate improvements in strict geometric accuracy and overall
reconstruction quality. Although the best predictor varies across
individual blocks, GeoFF3D provides the strongest average performance.
Figure~\ref{fig:qualitative_comparison} illustrates the different boundary and vertical error patterns
behind these metrics.

\begin{table}[t]
  \centering
  {
    \renewcommand{\arraystretch}{1.08}
    \setlength{\tabcolsep}{5.0pt}
    \begin{tabular}{@{}lcccc@{}}
      \toprule
      Method & Acc.$\downarrow$ & Comp.$\downarrow$ & F@1$\uparrow$ & F@5$\uparrow$ \\
      \midrule
      VGGT-SLAM & 21.87 & 32.03 & 0.060 & 0.323 \\
      VGGT-SLAM 2.0 & 10.36 & 36.15 & 0.108 & 0.411 \\
      VGGT-Long & 9.49 & 47.97 & 0.110 & 0.408 \\
      TTT3R & 10.48 & 79.50 & 0.061 & 0.239 \\
      LingBot-Map & 7.75 & 44.02 & 0.114 & 0.451 \\
      VGGT + SLRF & 8.00 & 5.21 & 0.125 & 0.608 \\
      Pi3X + SLRF & \underline{6.05} & \underline{4.40} & \underline{0.181} & \underline{0.687} \\
      GeoFF3D & \textbf{4.14} & \textbf{2.28} & \textbf{0.319} & \textbf{0.848} \\
      \bottomrule
    \end{tabular}
  }
  \caption{Average comparison on UAVScenes.}
  \label{tab:uavscenes_results}
\end{table}

\paragraph{Long UAV sequences.}
The margin becomes larger on UAVScenes: compared with Pi3X + SLRF,
GeoFF3D reduces Accuracy and Completeness from 6.05~m and 4.40~m to
4.14~m and 2.28~m, by 31.5\% and 48.2\%, respectively, while improving F@1 from 0.181
to 0.319 and F@5 from 0.687 to 0.848. The consistent gains at both
thresholds show improvements in fine-scale accuracy and overall
reconstruction quality. These sequences have elongated and
near-collinear camera layouts, for which full Sim(3) weakly constrains
roll and pitch. The quantitative gains and side profiles in
Fig.~\ref{fig:qualitative_comparison} therefore support the benefit of
establishing metric scale and gravity during prediction.


\subsection{Ablations and Analysis}\label{subsec:ablations}

Our ablations examine the coordinate-anchored model, the SLRF pipeline,
pose-prior robustness, chunk-size sensitivity, and efficiency and scalability.

\paragraph{Coordinate-anchored model.}
Table~\ref{tab:coordinate_anchor_ablation} evaluates 16-view samples from UseGeo and UAVFF3D-Real using ATE, Chamfer-L1 distance (CD), and gravity-direction error (GDE), defined as camera-center RMSE, symmetric point-cloud distance, and mean camera-up angular error, respectively.
Removing the world-frame losses causes the largest degradation, showing
that they establish the coordinate system defined by the translation anchors.
Gravity supervision consistently reduces GDE, showing that it further
stabilizes the Z-up orientation. For the full model, Sim(3) lowers ATE but
increases CD and GDE relative to GA-Sim, revealing a trade-off between fitting
camera centers and preserving gravity-aligned geometry.

\begin{table}[t]
  \centering
  \small
  \renewcommand{\arraystretch}{1.03}
  \setlength{\tabcolsep}{1.2pt}
  \begin{tabular}{@{}llccc|ccc@{}}
    \toprule
    \multirow{2}{*}{Variant}
    & \multirow{2}{*}{Align.}
    & \multicolumn{3}{c|}{\textit{UseGeo}}
    & \multicolumn{3}{c}{\textit{UAVFF3D-Real}} \\
    & & ATE$\downarrow$ & CD$\downarrow$ & GDE$\downarrow$
    & ATE$\downarrow$ & CD$\downarrow$ & GDE$\downarrow$ \\
    \midrule
    (a) Pi3X & Sim(3)
      & \textbf{1.18} & 1.67 & 1.11
      & \textbf{4.37} & \underline{8.95} & 2.40 \\
    (b) w/o world & GA-Sim
      & 3.13 & 3.37 & 4.04
      & 22.18 & 22.95 & 12.87 \\
    (c) w/o gravity & GA-Sim
      & 2.02 & \textbf{1.35} & 1.03
      & 5.85 & \textbf{8.91} & \underline{1.55} \\
      \midrule
    (d) Full & GA-Sim
      & 1.82 & \underline{1.62} & \underline{0.95}
      & 5.26 & 9.37 & \textbf{1.16} \\
    (e) Full & Sim(3)
      & \underline{1.38} & 2.13 & 2.02
      & \underline{4.61} & 15.14 & 2.71 \\
    (f) Full & None
      & 5.34 & 5.27 & \textbf{0.93}
      & 11.36 & 10.12 & \textbf{1.16} \\
    \bottomrule
  \end{tabular}
  \caption{Coordinate-anchored model ablation. ``w/o world'' removes
  $\mathcal L_{\mathrm{world}}$ and $\mathcal L_{\mathrm{pose}}$,
  while ``w/o gravity'' removes $\mathcal L_{\mathrm{grav}}$.}
  \label{tab:coordinate_anchor_ablation}
\end{table}

\paragraph{SLRF pipeline.}
The following full-pipeline analyses report averages over UseGeo D1,
UAVFF3D-Real NF01, and UAVScenes Town01. We report CD, ATE, 3D seam error
($S_{\mathrm{3D}}$), and vertical seam error ($S_z$). $S_{\mathrm{3D}}$
is the mean Euclidean discrepancy between corresponding points
reconstructed from shared views in neighboring chunks, while $S_z$ is
their mean absolute difference along the vertical axis.
Replacing the temporal-sequential pipeline (a) with the footprint-hierarchical pipeline (b) substantially improves global accuracy. Variant (c) achieves the lowest CD, while (d) provides the strongest seam consistency. The full model obtains the lowest ATE with competitive CD and seam errors.

\begin{table}[h]
  \centering
  \small
  \renewcommand{\arraystretch}{1.03}
  \setlength{\tabcolsep}{1.5pt}
  \begin{tabular}{@{}lcccc@{\hskip 3pt}cccc@{}}
    \toprule
    Variant & T & H & G & P
    & CD$\downarrow$ & ATE$\downarrow$
    & $S_{\mathrm{3D}}\downarrow$ & $S_z\downarrow$ \\
    \midrule
    (a) & \checkmark & & & &
      34.85 & 57.24 & 1.71 & 0.95 \\
    (b) & & \checkmark & & &
      7.55 & 8.27 & 1.38 & 0.86 \\
    (c) & & \checkmark & \checkmark & &
      \textbf{4.23} & 6.52 & 1.60 & 1.04 \\
    (d) & & \checkmark & & \checkmark &
      5.55 & \underline{6.27} & \textbf{1.03} & \textbf{0.57} \\
    Full & & \checkmark & \checkmark & \checkmark &
      \underline{4.50} & \textbf{5.52}
      & \underline{1.29} & \underline{0.79} \\
    \bottomrule
  \end{tabular}
  \caption{SLRF pipeline ablation. T and H denote the
    temporal-sequential and footprint-hierarchical pipelines,
    respectively; G replaces Sim(3) with GA-Sim and GA-Rigid,
    and P adds prior propagation.}
  \label{tab:slrf_ablation}
\end{table}

\paragraph{Pose-prior robustness.}
Table~\ref{tab:pose_prior_robustness} evaluates doubled pose noise
and the removal of rotation priors.
GeoFF3D remains relatively stable under doubled noise, whereas using translation priors alone causes substantially larger degradation in both global accuracy and seam consistency.

\begin{table}[h]
  \centering
  \small
  \renewcommand{\arraystretch}{1.03}
  \setlength{\tabcolsep}{3.2pt}
  \begin{tabular}{@{}lcccc@{}}
    \toprule
    Setting
    & CD$\downarrow$
    & ATE$\downarrow$
    & $S_{\mathrm{3D}}\downarrow$
    & $S_z\downarrow$ \\
    \midrule
    Default noise
      & \textbf{4.50} & \textbf{5.52} & \textbf{1.29} & \textbf{0.79} \\
    $2\times$ pose noise
      & 5.12 & 6.34 & 1.73 & 1.21 \\
    Translation only
      & 6.56 & 10.67 & 2.48 & 1.76 \\
    \bottomrule
  \end{tabular}
  \caption{Robustness to pose-prior.}
  \label{tab:pose_prior_robustness}
\end{table}

\paragraph{Chunk-size sensitivity.}
Table~\ref{tab:chunk_size_sensitivity} evaluates chunk budgets of
20, 30, and 40 views. The 20-view setting is fastest and uses the least memory, while $M=30$ achieves the best CD and ATE with moderate resource use. Increasing the budget to 40 views raises peak memory and degrades both global metrics. We therefore use $M=30$ to balance reconstruction quality and efficiency.

\begin{table}[t]
  \centering
  \small
  \renewcommand{\arraystretch}{1.03}
  \setlength{\tabcolsep}{3.6pt}
  \begin{tabular}{@{}ccccc@{}}
    \toprule
    $M$
    & CD$\downarrow$
    & ATE$\downarrow$
    & Avg. Time (s)$\downarrow$
    & Peak Mem. (GiB)$\downarrow$ \\
    \midrule
    20 & 4.56 & 6.47
       & \textbf{70.59} & \textbf{11.82} \\
    30 & \textbf{4.50} & \textbf{5.52}
       & 86.44 & 12.59 \\
    40 & 5.15 & 6.05
       & 77.86 & 16.50 \\
    \bottomrule
  \end{tabular}
  \caption{Chunk-size sensitivity. CD, ATE, and time are averaged over the three scenes; peak memory is their maximum.}
  \label{tab:chunk_size_sensitivity}
\end{table}

\begin{figure}[h]
  \centering
  \includegraphics[width=\columnwidth]
    {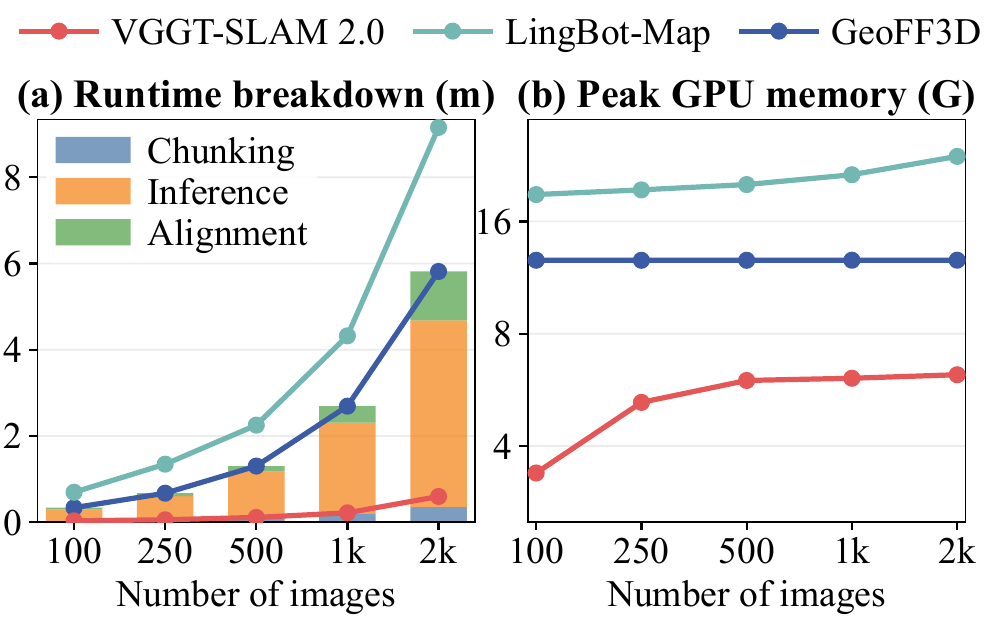}
  \caption{Runtime breakdown and peak GPU memory versus image count on UAVScenes Town01.}
  \label{fig:efficiency_scalability}
\end{figure}

\paragraph{Efficiency and scalability.}
Figure~\ref{fig:efficiency_scalability} evaluates the complete system
as the image count increases. Runtime grows nearly linearly, with model inference accounting for most of the cost. GeoFF3D reconstructs 2,000 images in about five minutes using roughly 16~GiB of GPU memory.


\section{Conclusions}\label{sec:conclusions}

We presented GeoFF3D, which directly reconstructs cameras and dense geometry in a georeferenced,
gravity-aligned Z-up metric frame. Experiments show that GeoFF3D achieves the best average performance on aerial mapping
blocks and a larger advantage on long, near-collinear sequences, while outperforming representative SLAM and
streaming systems. The results show that coordinate anchoring benefits general multi-strip aggregation and
that metric scale and gravity are especially important for elongated trajectories. Future work will explore online reconstruction, global optimization, and improved robustness to sparse or noisy georeferencing priors.

\bibliography{refs}

\clearpage
\appendix
\onecolumn
\pdfdest name {supp-quantitative} xyz
\section{Per-Sequence Quantitative Results}
\label{supp:quantitative}

The main paper reports averages over UAVScenes. Here we expose the
sequence-level variation hidden by those averages, separating directed
distance errors from thresholded F-scores.

\pdfdest name {supp-sequence-results} xyz
\subsection{UAVScenes}
\label{supp:sequence_results}

Tables~\ref{tab:supp_uavscenes_distance} and
\ref{tab:supp_uavscenes_fscore} use the main-paper configuration of each
method: VGGT-SLAM is
evaluated with Sim(3) alignment, all SLRF methods use 30-view chunks, and
Pi3X + \framework{} and \method{} receive the same noisy pose priors.

\begin{center}
\begin{minipage}{\textwidth}
\centering
{\small\setlength{\tabcolsep}{2.4pt}\renewcommand{\arraystretch}{0.96}
\begin{tabular}{@{}lcccccccc@{}}
\toprule
\multicolumn{9}{c}{\textit{(a) Accuracy (m) $\downarrow$}} \\
\midrule
Method & Town01 & Town02 & Valley01 & Valley02 & Airport01 & Airport02 & Island01 & Island02 \\
\midrule
VGGT-SLAM (Sim(3)) & 29.991 & 27.393 & 30.978 & 28.674 & 6.828 & 9.136 & 29.864 & 12.119 \\
VGGT-SLAM 2.0 & 5.789 & 22.608 & \best{6.696} & 19.519 & 3.400 & 6.413 & 5.616 & 12.848 \\
VGGT-Long & 4.083 & 7.320 & 17.774 & 19.611 & 2.715 & 5.680 & 9.825 & 8.874 \\
TTT3R & 21.071 & 10.449 & 13.199 & \best{7.409} & 5.854 & 9.598 & 8.374 & 7.851 \\
LingBot-Map & 5.069 & 3.619 & 11.056 & 16.154 & 4.949 & 5.476 & 7.806 & 7.859 \\
VGGT + \framework & 9.542 & 10.555 & 10.908 & 11.162 & 3.287 & 6.110 & 6.294 & 6.148 \\
Pi3X + \framework & 11.221 & 6.249 & 9.645 & 7.543 & 2.883 & 3.177 & 3.939 & 3.712 \\
\method & \best{2.112} & \best{1.966} & 12.712 & 8.646 & \best{1.710} & \best{1.579} & \best{2.303} & \best{2.092} \\
\midrule
\multicolumn{9}{c}{\textit{(b) Completeness (m) $\downarrow$}} \\
\midrule
Method & Town01 & Town02 & Valley01 & Valley02 & Airport01 & Airport02 & Island01 & Island02 \\
\midrule
VGGT-SLAM (Sim(3)) & 70.925 & 96.866 & 30.536 & 23.604 & 4.949 & 4.204 & 19.626 & 5.528 \\
VGGT-SLAM 2.0 & 90.787 & 18.684 & 148.922 & 8.863 & 3.232 & 3.094 & 5.397 & 10.260 \\
VGGT-Long & 188.463 & 113.174 & 14.829 & 49.660 & 3.308 & 4.072 & 5.565 & 4.721 \\
TTT3R & 59.267 & 53.307 & 125.724 & 124.355 & 31.376 & 24.836 & 102.990 & 114.123 \\
LingBot-Map & 36.116 & 109.232 & 133.475 & 59.554 & 4.876 & 4.402 & 2.128 & 2.387 \\
VGGT + \framework & 8.525 & 10.886 & 6.078 & 7.311 & 2.154 & 2.435 & 2.411 & 1.872 \\
Pi3X + \framework & 8.510 & 6.627 & 6.977 & 4.902 & 2.281 & 2.330 & 1.971 & 1.625 \\
\method & \best{1.746} & \best{1.580} & \best{4.432} & \best{4.660} & \best{1.527} & \best{1.252} & \best{1.663} & \best{1.398} \\
\bottomrule
\end{tabular}}
\captionof{table}{Per-sequence Accuracy and Completeness on UAVScenes.
Both metrics are measured in meters. Best results are shown in bold.}
\label{tab:supp_uavscenes_distance}
\end{minipage}
\end{center}

\begin{center}
\begin{minipage}{\textwidth}
\centering
{\small\setlength{\tabcolsep}{2.4pt}\renewcommand{\arraystretch}{0.96}
\begin{tabular}{@{}lcccccccc@{}}
\toprule
\multicolumn{9}{c}{\textit{(a) F@1 $\uparrow$}} \\
\midrule
Method & Town01 & Town02 & Valley01 & Valley02 & Airport01 & Airport02 & Island01 & Island02 \\
\midrule
VGGT-SLAM (Sim(3)) & 0.019 & 0.021 & 0.041 & 0.033 & 0.106 & 0.105 & 0.040 & 0.116 \\
VGGT-SLAM 2.0 & 0.001 & 0.032 & 0.003 & 0.061 & 0.270 & 0.177 & 0.193 & 0.123 \\
VGGT-Long & 0.008 & 0.023 & 0.030 & 0.045 & 0.306 & 0.211 & 0.126 & 0.130 \\
TTT3R & 0.023 & 0.052 & 0.037 & 0.075 & 0.135 & 0.104 & 0.035 & 0.024 \\
LingBot-Map & 0.094 & 0.048 & 0.034 & 0.051 & 0.165 & 0.182 & 0.154 & 0.182 \\
VGGT + \framework & 0.062 & 0.105 & 0.056 & 0.067 & 0.204 & 0.186 & 0.154 & 0.170 \\
Pi3X + \framework & 0.065 & 0.095 & \best{0.083} & \best{0.105} & 0.257 & 0.305 & 0.228 & 0.313 \\
\method & \best{0.284} & \best{0.297} & 0.055 & 0.078 & \best{0.412} & \best{0.465} & \best{0.453} & \best{0.510} \\
\midrule
\multicolumn{9}{c}{\textit{(b) F@5 $\uparrow$}} \\
\midrule
Method & Town01 & Town02 & Valley01 & Valley02 & Airport01 & Airport02 & Island01 & Island02 \\
\midrule
VGGT-SLAM (Sim(3)) & 0.149 & 0.142 & 0.247 & 0.240 & 0.555 & 0.541 & 0.220 & 0.488 \\
VGGT-SLAM 2.0 & 0.019 & 0.241 & 0.025 & 0.390 & 0.806 & 0.681 & 0.677 & 0.452 \\
VGGT-Long & 0.034 & 0.096 & 0.260 & 0.263 & 0.823 & 0.711 & 0.539 & 0.537 \\
TTT3R & 0.202 & 0.372 & 0.125 & 0.192 & 0.431 & 0.446 & 0.085 & 0.057 \\
LingBot-Map & 0.490 & 0.134 & 0.109 & 0.275 & 0.679 & 0.682 & 0.608 & 0.635 \\
VGGT + \framework & 0.450 & 0.596 & 0.448 & 0.456 & 0.850 & 0.705 & 0.682 & 0.677 \\
Pi3X + \framework & 0.446 & 0.586 & 0.489 & 0.560 & 0.867 & 0.848 & 0.835 & 0.862 \\
\method & \best{0.937} & \best{0.966} & \best{0.492} & \best{0.578} & \best{0.960} & \best{0.976} & \best{0.925} & \best{0.950} \\
\bottomrule
\end{tabular}}
\captionof{table}{Per-sequence F-scores on UAVScenes at 1\,m and 5\,m
thresholds. Best results are shown in bold.}
\label{tab:supp_uavscenes_fscore}
\end{minipage}
\end{center}

\clearpage
\raggedbottom
\pdfdest name {supp-qualitative} xyz
\section{Qualitative Results}
\label{supp:qualitative}

The qualitative comparisons complement the aggregate metrics by exposing
coverage, boundary continuity, vertical orientation, and cross-dataset
generalization. We organize the examples by aerial blocks, long trajectories,
and NPU-DroneMap sequences.

\pdfdest name {supp-aerial-qualitative} xyz
\subsection{Aerial Mapping Blocks}
\label{supp:aerial_qualitative}

Figures~\ref{fig:supp_usegeo} and \ref{fig:supp_uavff3d} use matched cameras
within each row so that coverage, boundary completeness, local deformation,
and missing regions can be compared directly. The trajectory column also makes
the acquisition geometry explicit, which is essential when interpreting
difficult near-collinear or sparsely overlapping blocks.

\begin{center}
\begin{minipage}{\textwidth}
\centering
\includegraphics[width=\textwidth]{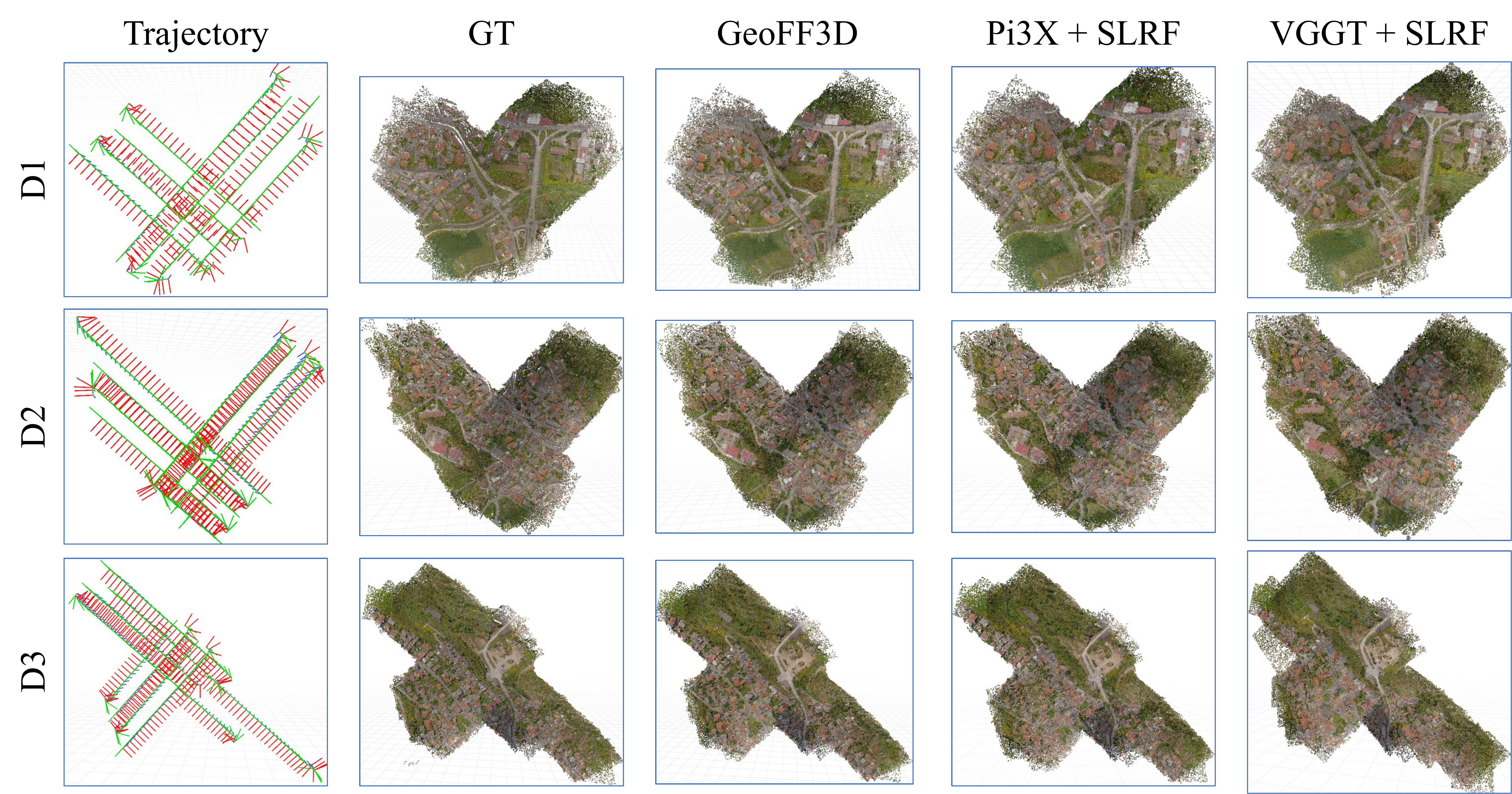}
\captionof{figure}{Matched-view qualitative comparison on all three UseGeo blocks. Rows
correspond to D1--D3; columns show the camera trajectory, ground truth,
\method{}, Pi3X + \framework, and VGGT + \framework. All reconstructions in a
row use the same rendering camera and bounds.}
\label{fig:supp_usegeo}
\end{minipage}
\end{center}

\begin{center}
\begin{minipage}{\textwidth}
\centering
\includegraphics[width=\textwidth]{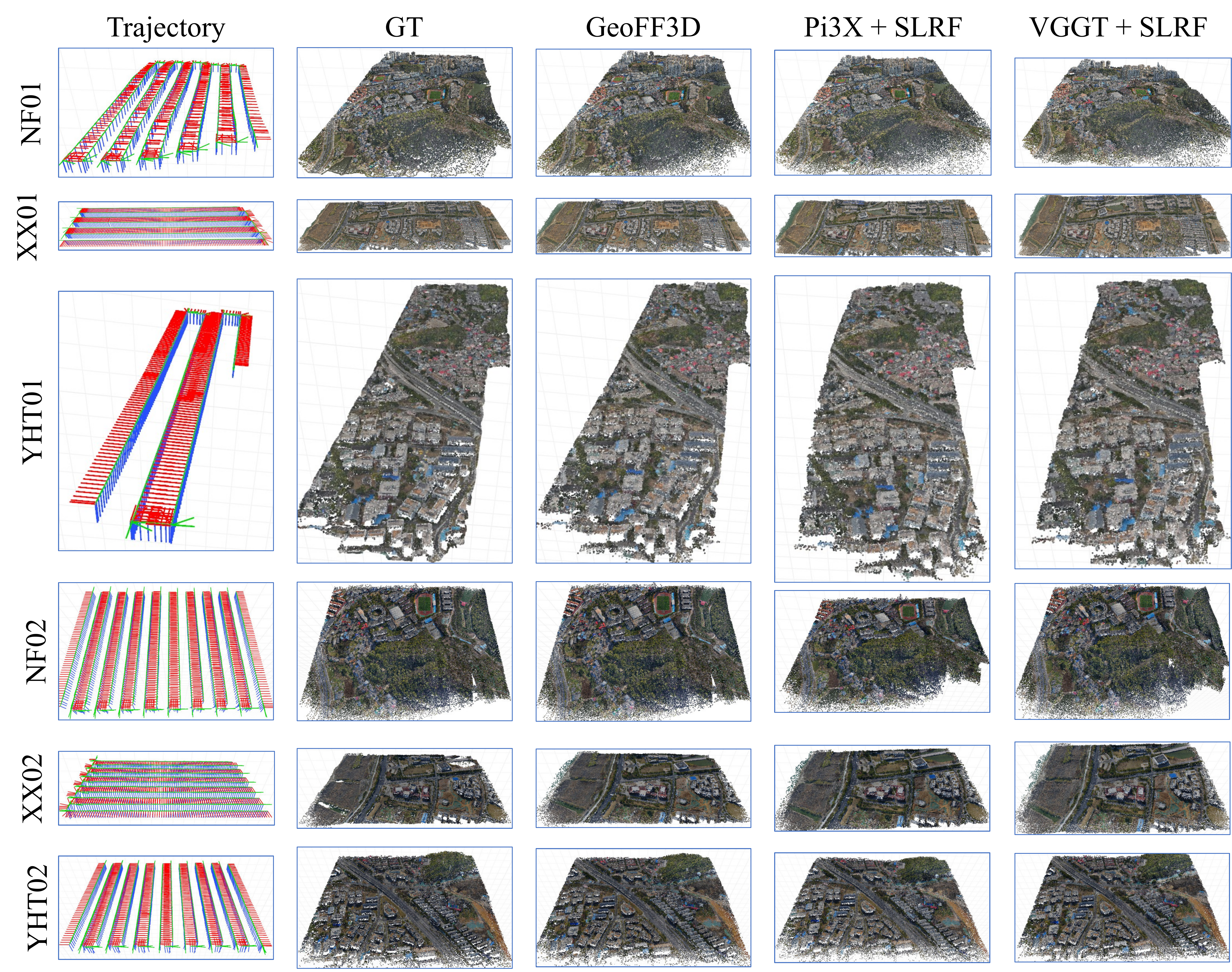}
\captionof{figure}{Matched-view qualitative comparison on five UAVFF3D-Real blocks.
Rows show NF01, XX01, YHT01, NF02, and XX02; columns show the trajectory, ground
truth, \method{}, Pi3X + \framework, and VGGT + \framework. The same view and
spatial bounds are used within each row.}
\label{fig:supp_uavff3d}
\end{minipage}
\end{center}

\pdfdest name {supp-long-qualitative} xyz
\subsection{Long UAV Sequences}
\label{supp:long_qualitative}

Near-collinear long trajectories are particularly sensitive to post-hoc roll
and pitch estimation. The qualitative comparison therefore reports both a top
view and a vertical side profile. A method may fit the horizontal camera layout
while producing an incorrect vertical orientation or excessive scene
thickness. \method{} retains a Z-up profile because gravity is constrained during
prediction and preserved during hierarchical aggregation.

\begin{center}
\begin{minipage}{\textwidth}
\centering
\includegraphics[width=\textwidth]{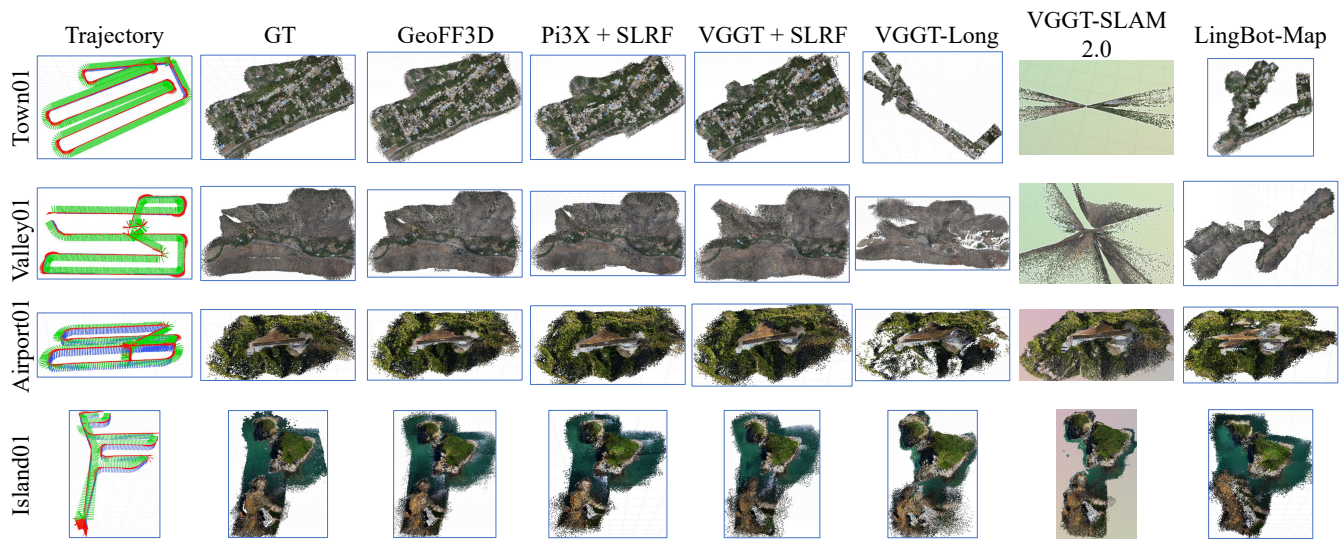}
\captionof{figure}{Matched-view comparison on representative UAVScenes trajectories.
Rows show Town01, Valley01, Airport01, and Island01; columns show the
trajectory, ground truth, \method{}, three chunked/streaming baselines, and
LingBot-Map. The shared views expose fragmentation and global orientation
errors that are obscured by aggregate metrics.}
\label{fig:supp_uavscenes}
\end{minipage}
\end{center}

\pdfdest name {supp-npu-generalization} xyz
\subsection{NPU-DroneMap Generalization}
\label{supp:npu_generalization}

NPU-DroneMap contains real keyframe sequences spanning urban, industrial, road,
and agricultural environments. Although complete dense reference point clouds
are unavailable, these sequences test whether the coordinate-anchored model
and spatial pipeline generalize beyond the quantitative benchmark datasets.

\begin{center}
\begin{minipage}{\textwidth}
\centering
\includegraphics[width=\textwidth]{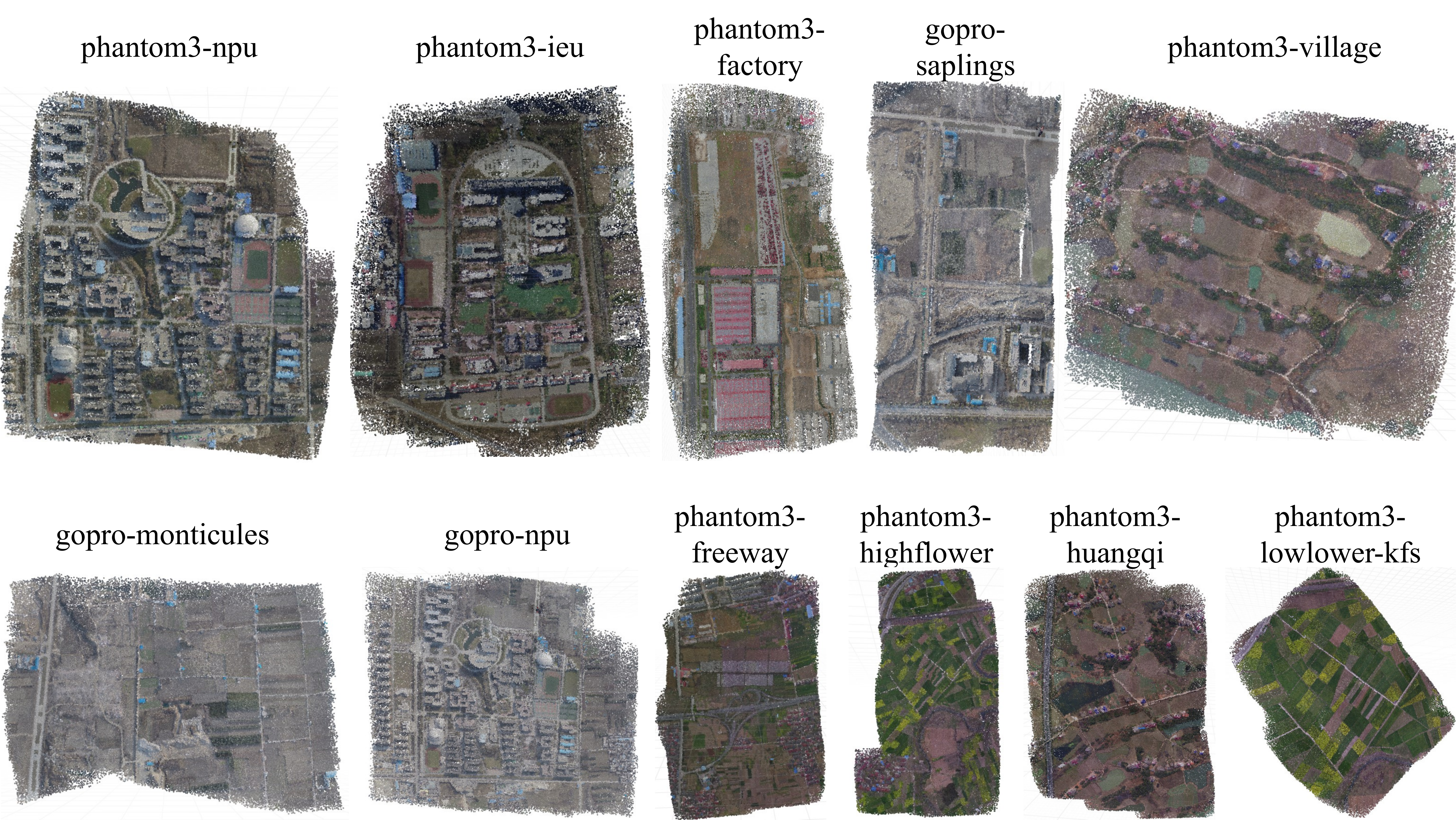}
\captionof{figure}{Cross-dataset generalization on 11 representative NPU-DroneMap
keyframe sequences (from the 12-sequence dataset), covering campus, factory,
road, industrial, village, and agricultural scenes. All reconstructions use a
common point-rendering convention; no dense reference geometry is available
for quantitative scoring.}
\label{fig:supp_npu}
\end{minipage}
\end{center}

\end{document}